\documentclass{article}

    \PassOptionsToPackage{numbers, compress}{natbib}
    \usepackage[final]{neurips_2024}

\usepackage[utf8]{inputenc} % allow utf-8 input
\usepackage[T1]{fontenc}    % use 8-bit T1 fonts
\usepackage{hyperref}       % hyperlinks
\usepackage{url}            % simple URL typesetting
\usepackage{booktabs}       % professional-quality tables
\usepackage{amsfonts}       % blackboard math symbols
\usepackage{nicefrac}       % compact symbols for 1/2, etc.
\usepackage{microtype}      % microtypography
\usepackage{xcolor}         % colors
\usepackage{bbding}       

\usepackage{algorithm}
\usepackage{algpseudocodex}
\usepackage{amsmath}
\usepackage{amssymb}
\usepackage{multirow}
\usepackage{makecell}
\usepackage{subcaption}
\usepackage{fancybox}
\usepackage{fancyvrb}
\usepackage{bm}
\usepackage{enumerate}

\DeclareMathOperator*{\argmin}{argmin}

\newcommand{\name}{SocialGPT}
\newcommand{\namep}{SocialPrompt}
\definecolor{OliveGreen}{rgb}{0,0.6,0}
\definecolor{ForestGreen}{rgb}{0.133,0.545,0.133}
\definecolor{LavaRed}{rgb}{0.800,0.200,0.200}

\title{SocialGPT: Prompting LLMs for Social Relation Reasoning via Greedy Segment Optimization}

\author{Wanhua Li$^{\ast,1}$ Zibin Meng$^{\ast,1,2}$ Jiawei Zhou$^{3}$ Donglai Wei$^{4}$ Chuang Gan$^{5,6}$ Hanspeter Pfister$^{1}$\\[1mm]
$^{1}$Harvard University\quad $^{2}$Tsinghua University\quad$^{3}$Stony Brook University\\
$^{4}$Boston College\quad$^{5}$MIT-IBM Watson AI Lab\quad$^{6}$UMass Amherst
}

\begin{document}

\maketitle
{\let\thefootnote\relax\footnotetext{{$^{\ast}$ Equal contribution.}}}

\begin{abstract}
Social relation reasoning aims to identify relation categories such as friends, spouses, and colleagues from images. While current methods adopt the paradigm of training a dedicated network end-to-end using labeled image data, they are limited in terms of generalizability and interpretability. To address these issues, we first present a simple yet well-crafted framework named {\name}, which combines the perception capability of Vision Foundation Models (VFMs) and the reasoning capability of Large Language Models (LLMs) within a modular framework, providing a strong baseline for social relation recognition. Specifically, we instruct VFMs to translate image content into a textual social story, and then utilize LLMs for text-based reasoning. {\name} introduces systematic design principles to adapt VFMs and LLMs separately and bridge their gaps. Without additional model training, it achieves competitive zero-shot results on two databases while offering interpretable answers, as LLMs can generate language-based explanations for the decisions. The manual prompt design process for LLMs at the reasoning phase is tedious and an automated prompt optimization method is desired. As we essentially convert a visual classification task into a generative task of LLMs, automatic prompt optimization encounters a unique long prompt optimization issue. To address this issue, we further propose the Greedy Segment Prompt Optimization (GSPO), which performs a greedy search by utilizing gradient information at the segment level. Experimental results show that GSPO significantly improves performance, and our method also generalizes to different image styles. The code is available at \url{https://github.com/Mengzibin/SocialGPT}.
\end{abstract}

\section{Introduction}
\label{sec:intro}

Social relationships are of paramount importance in our lives, as they significantly impact our emotional, psychological, and physical well-being. Social relationship recognition aims to categorize the relationships such as friends, colleagues, band members, and so on, that exist between individuals given an input image and the bounding boxes of the two persons of interest~\cite{sun2017domain}. 
In recent years, social relationship recognition has garnered significant attention~\cite{sun2017domain,zhang2019multi,li2020graph,wang2023shifted} due to its wide range of applications, including product recommendation~\cite{park2009individual}, autonomous systems~\cite{wang2018deep}, and more. 

\begin{figure}[t]
\centering
\includegraphics[width=1.0\linewidth]{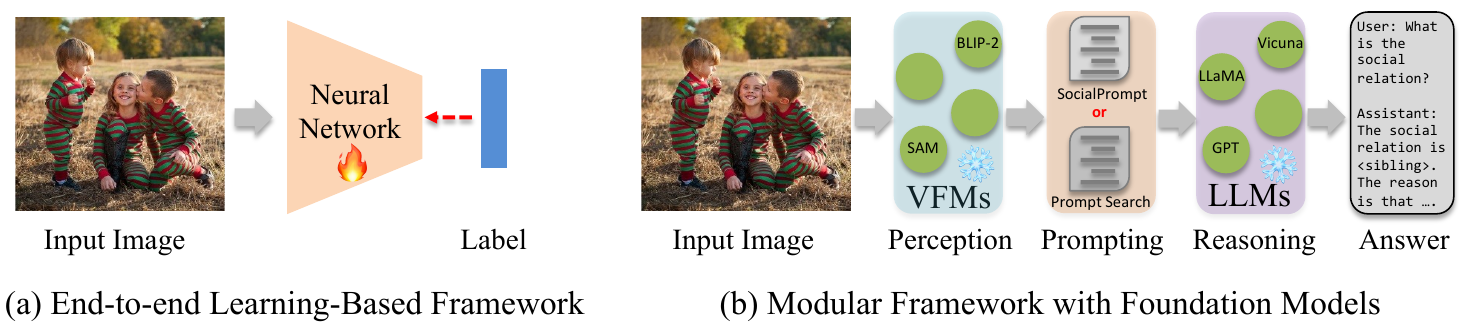}
% \vspace{-1in}
\caption{(a) End-to-end learning-based framework for social relation reasoning. A dedicated neural network is trained end-to-end with full training data. (b) We propose a modular framework with foundation models for social relation reasoning. Our proposed {\name} first employs VFMs to extract visual information into textual format,
and then perform text-based reasoning with LLMs, using either our manually designed SocialPrompt or optimized prompts.}
\label{fig:teaser}
\vspace{-16pt}
\end{figure}

Over the past decade, the field of computer vision has witnessed tremendous success~\cite{ren2015faster,he2016deep,he2017mask,li2019bridgenet,strudel2021segmenter,carion2020end} in the end-to-end learning framework, 
which trains a dedicated neural network end-to-end on a customized dataset. 
Research in social relationship recognition has also followed a similar trajectory~\cite{sun2017domain,li2017dual,zhang2019multi}. 
As social relationship reasoning represents a cognitive function that operates at a higher level than visual perception, many methods~\cite{wang2018deep,li2020graph} incorporate rich prior knowledge of social relations into the models. For example, GRM~\cite{wang2018deep} integrated a knowledge graph into its model to leverage the information of contextual objects. GR$^2$N~\cite{li2020graph} and TRGAT~\cite{guo2023social} exploit the logical constraints among multiple social relationships within the same scene. While these methods have achieved notable results, they are limited in terms of generalization and interpretability. In other words, we cannot trust that the trained models can generalize to arbitrary scenarios, and these models fail to provide the reasons and explanations for their decisions. 

In this paper, we first present a modular framework with foundation models for social relation reasoning. Recently, we have witnessed the significant success of foundational models~\cite{bommasani2021opportunities}. Many Vision Foundation Models (VFMs) can accurately perform basic visual perception tasks such as identifying ``what'' and ``where'' in images~\cite{li2023blip,kirillov2023segment,qin2024langsplat,radford2021learning}. On the other hand, the emergence of Large Language Models (LLMs) demonstrates strong reasoning capabilities~\cite{wei2022chain,wei2022emergent,wang2022self,yao2023tree}. Therefore, we present a framework that follows the ``perceive with VFMs, reason with LLMs'' paradigm. This framework first employs VFMs to convert images into textual data, and subsequently leverages the textual reasoning capabilities of LLMs for relation prediction. In this process, VFMs process visual signals into fundamental facts, and then LLMs analyze these facts to make explainable inferences.

Our framework performs visual reasoning for \textbf{Social} relationship recognition using \textbf{GPT}-style LLMs, coined {\name}.
{\name} introduces systematic design principles to guide and adapt VFMs and LLMs for social relationship reasoning.
Specifically, in the perception phase, we extract both comprehensive and domain-specific visual information with VFMs, which is further fused into a coherent textual social story with symbol-based object reference and is easily readable.
In the reasoning phase, we utilize a structured social relation reasoning prompt, named {\namep}, composed of different segments for ``system, expectation, context, and guidance'' to better instruct LLMs.
With the proposed systematic design principles, our {\name} provides a strong baseline and achieves highly competitive zero-shot results, compared to the state-of-the-art methods that undergo end-to-end training on full training datasets.

Lastly, we observed that LLMs exhibit high sensitivity to prompts during the reasoning process, but the manual prompt design is a time-consuming and labor-intensive task~\cite{liu2023pre,shin2020autoprompt}. 
We propose the Greedy Segment Prompt Optimization (GSPO) algorithm for automatic prompt tuning. As we convert a visual classification task as a generative task of LLMs, automatic prompt tuning for {\namep} encounters the long prompt optimization issue. Our proposed GSPO addresses these issues by utilizing gradient information at the segment level for greedy search.
Experiments demonstrate that GSPO significantly improves the performance of LLMs.
Figure~\ref{fig:teaser} visualizes our paradigm. 
To summarize, we make the following contributions:
1). We present a simple modular framework with foundation models for social relation reasoning, which provides a strong baseline as the first zero-shot social relation recognition method.
2). To address the long prompt optimization issue associated with visual reasoning tasks, we further propose the Greedy Segment Prompt Optimization, which performs a greedy search on the segment level with gradient guidance.
3). Experiments demonstrate that our method attains very competitive and explainable zero-shot results without additional model training. With GSPO, our method significantly outperforms the state-of-the-art methods.

\section{Related Work}
\label{sec:related}

\textbf{Foundation Models.} 
Recently, we have witnessed the tremendous success of foundational models~\cite{radford2021learning,rombach2022high,openai2023gpt4,li2022ordinalclip,vicuna2023}. 
Foundation models are typically trained on massive data, possess a large number of model parameters, and exhibit excellent performance along with strong generalization capabilities~\cite{bommasani2021opportunities}. 
The emergence of LLMs~\cite{bommasani2021opportunities,openai2023gpt4,zhang2022opt,anil2023palm} has significantly reshaped the field of Natural Language Processing (NLP). ChatGPT and GPT-4~\cite{openai2023gpt4}, developed by OpenAI, are among the most famous LLMs. GPT-4, in particular, demonstrates a strikingly close-to-human-level intelligence~\cite{bubeck2023sparks}. Meanwhile, many open-source LLMs like Vicuna~\cite{vicuna2023}, LLaMa~\cite{touvron2023llama}, and LLaMa-2~\cite{touvron2023llama2} have been developed, and have achieved outstanding performance across various NLP tasks. On the other hand, VFMs~\cite{radford2021learning,rombach2022high,caron2021emerging,oquab2023dinov2,li2022grounded,ramesh2021zero} have also made significant advancements. CLIP~\cite{radford2021learning} connects images and text, enabling zero-shot image classification~\cite{zhou2022learning,menon2022visual}. BLIP~\cite{li2022blip} and BLIP-2~\cite{li2023blip} demonstrate strong zero-shot image-to-text generation capabilities. SAM~\cite{kirillov2023segment} offers a foundation model for image segmentation~\cite{ling2022ocp}. While foundation model-based frameworks have been proposed for many other tasks including few-shot visual recognition~\cite{zhang2023prompt,gao2023clip,zhou2022conditional}, visual question answering ~\cite{shao2023prompting,guo2023images,surismenon2023vipergpt}, and semantic segmentation~\cite{yu2023foundation}, 
our {\name} explicitly employs text as the bridge between VFMs and LLMs and then proposes symbol-based referencing to support unambiguous text queries.

\noindent\textbf{Social Relation Recognition.} Social psychologists have conducted extensive research on social relationships over decades~\cite{cohen2004social,conte1981circumplex}, resulting in several different social theories~\cite{bugental2000acquisition,fiske1992four}.
Sun \emph{et al.} ~\cite{sun2017domain} followed Bugental’s domain-based theory~\cite{bugental2000acquisition} and annotated the PIPA dataset, which has become one of the most popular benchmarks for social relation recognition. 
Li \emph{et al.} ~\cite{li2017dual} adopted the relational models theory~\cite{fiske1992four} and contributed the People in Social Context (PISC) dataset. 
A dual-glance model was further proposed to leverage multiple contextual regions.
With the well-established benchmarks, numerous end-to-end methods~\cite{goel2019end,li2020graph,guo2023social,zhang2019multi} have been proposed, effectively advancing the field of social relationship recognition. Some methods~\cite{goel2019end,wang2018deep} employed knowledge graphs to exploit scene and global contextual cues. Noticing that there usually are multiple social relations on the same image, Li \emph{et al.} ~\cite{li2020graph} proposed GR$^2$N to jointly infer all relations on an image with graph neural networks. TRGAT~\cite{guo2023social} further considered higher-order constraints for social relations on an image and achieved better results. These methods adopted the end-to-end learning-based paradigm, whereas we propose a modular framework with foundation models.

\section{{\name}}
\label{sec:method}

\begin{figure}[t]
    \centering
    \includegraphics[width=1.0\textwidth]{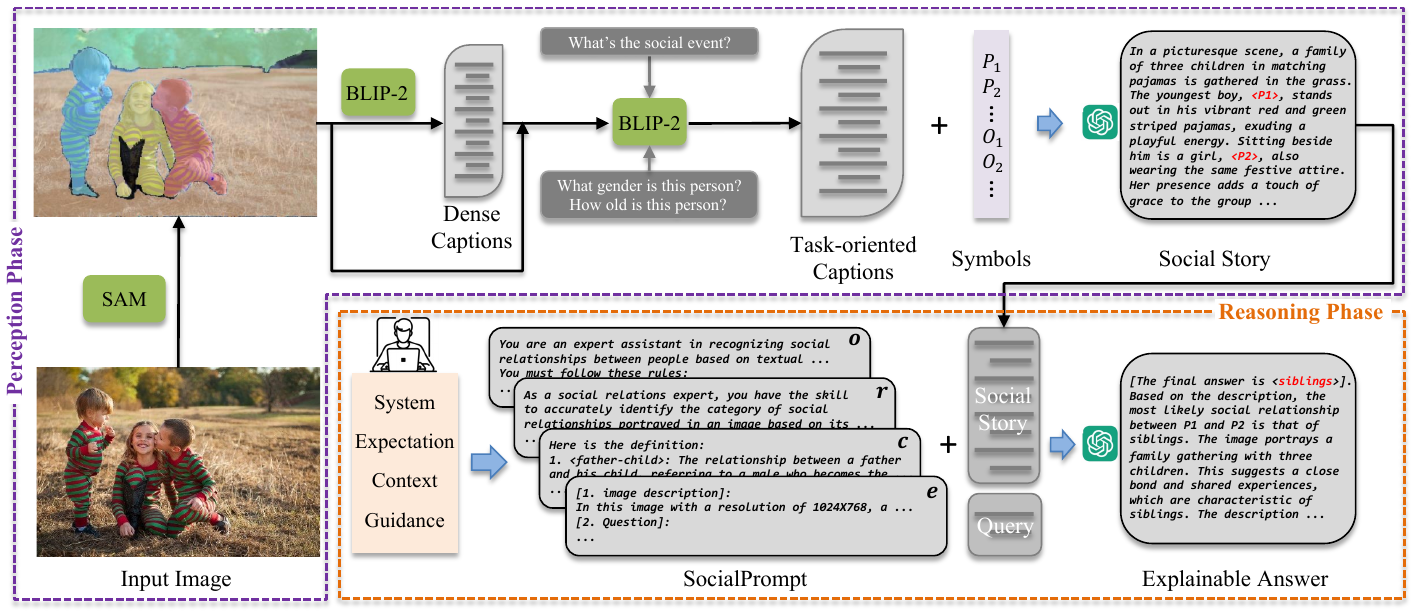}
    \caption{ The framework of {\name}, which follows the ``perception with VFMs, reasoning with LLMs'' paradigm. {\name} converts an image into a \textit{social story} in the perception phase, and then employs LLMs to generate explainable answers in the reasoning phase with  {\namep}.
    }  \label{fig:framework}
    \vspace{-16pt}
\end{figure}

Social relation recognition takes an image $\bm{I}$ and two bounding boxes $\bm{b}_1$ and $\bm{b}_2$ of two interested individuals as inputs, and requires a model that outputs the social relationship $\bm{y}$. 
We first introduce a modular framework with foundation models for social relation recognition in this section, which provides a strong zero-shot baseline. The pipeline is illustrated in Figure~\ref{fig:framework}.
On a high level, we first use VFMs to extract visual information at different granularities. The raw information is then fused into a coherent \textit{social story} in textual format, denoted as $\bm{S}$, which can be best reasoned with LLMs.

\subsection{Perception with Vision Foundation Models}

The perception objective is to extract essential visual information related to social relation reasoning, in order to connect with text-based LLMs for downstream reasoning. One straightforward approach is to utilize existing image captioning foundation models such as BLIP-2~\cite{li2023blip} to generate a caption or GPT-4V ~\cite{white2023chatgpt} to generate an image description. However, a single sentence or general-purpose description may overlook crucial details relevant to social relations present in the images.

We construct text-based visual information with VFMs with being both \textbf{comprehensive} and \textbf{domain-specific} as our guidelines.
To achieve this, we resort to the state-of-the-art image segmentation tool, the Segment Anything Model (SAM)~\cite{kirillov2023segment}, and the powerful vision-language foundation model, BLIP-2~\cite{li2023blip}, for both identifying important details in the image and describing them in language.
In particular, we use SAM to segment the image to obtain all different object masks, and then send individual objects by masking out others to BLIP-2 to obtain descriptions of each object. Together with the image-level caption, we formulate the \textit{dense captions} covering all objects in the input image.

The above gives us a comprehensive description of the image details. However, holistic captions of the image and different objects are not tailored to our task of social relation reasoning. To compensate for the lack of domain-specific information, we ask specific questions related to social identities by using the BLIP-2 dialog functionality to extract more specific information depending on object types. 
Recent research ~\cite{goel2019end,sun2017domain} has shown that the age and gender of individuals, as well as the social scene and activity, are important clues. 
Therefore, we actively inquire BLIP-2 about these clues. 
Specifically, when dealing with people objects, we inquire about age and gender details. This information is crucial for distinguishing familial relationships within a family unit, such as father-child and grandmother-grandchild relationships. For image-level captions, we explore the social scenario or event depicted in the picture. This approach allows us to generate \textit{task-oriented captions} that are tailored to our social relation recognition objective.

\subsection{Social Story Generation}

One could directly input the dense captions and task-oriented captions along with object axes and dimensions into LLMs for social relation reasoning, but the information is fragmented and objects are described in isolation. On the other hand, LLMs perform the best when working with human-readable natural language and they often struggle with arithmetic reasoning tasks ~\cite{imani2023mathprompter,patel2021nlp,schick2023toolformer}. Therefore, we integrate the aforementioned vision information by composing a social story that is complete and coherent. Objects are conveniently \textbf{referable} and described in relative relations, and the full story is easily \textbf{readable} by both humans and LLMs. This will serve as a crucial bridge from visual perception to textual reasoning, providing a solid foundation for the next step of understanding with LLMs.

We propose \textit{symbol-based referencing} for object referral. Multiple individuals and various social relationships coexist in a single image, and bounding boxes $\bm{b}_1$ and $\bm{b}_2$ are provided for specific relation inquiries in supervised learning settings. However, as we now convert the entire image into textual data and rely on LLMs for analysis, effective referral of individual objects becomes a critical question. Based on SAM segmentation masks, we can naturally derive bounding boxes for each object $i$ as $\bm{b}_i = [x_i, y_i, h_i, w_i]$, where $(x_i, y_i)$ is the center coordinate and $(h_i, w_i)$ are the height and width. While directly using these coordinates for referrals in the social story and question inquiries is precise, they pose extra challenges for readability and numerical reasoning for LLMs. Instead, we assign \textit{symbols} to each object to associate with its coordinates in the original image, textual caption, and task-specific features for our social story generation. We use $P_i$ to refer to people objects, and $O_i$ to refer to other objects. Numerical coordinates will not appear in our social story, and relative positional relations are described with the referral symbols. The symbol-based referring also enables straightforward querying for LLMs. For instance, one can directly inquire LLMs about the social relationship between $P_2$ and $P_3$ with natural language and LLMs will easily identify the queried persons associated with symbols. This provides a clear and concise bridge between the object descriptions and the bounding box-based queries, and a similar method can be adopted for a broader range of applications when text-based reasoning is involved for object referral for visual question answering, robotics, etc.

\begin{figure}[t]
\centering
\includegraphics[width=1.0\linewidth]{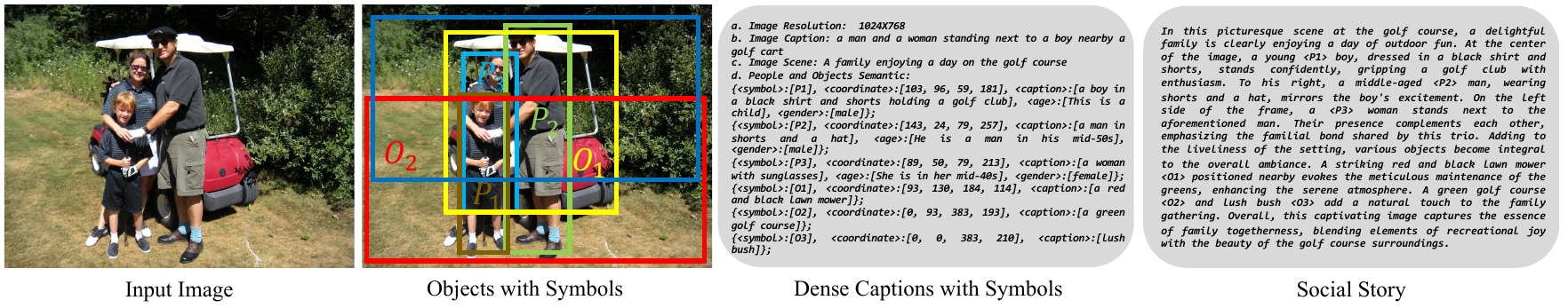}
\caption{An example of social story generation.}
% \vspace{-2mm}
\label{fig:dense_caption}
% \vspace{-16pt}
\end{figure}

Finally, based on the list of isolated image and object descriptions after symbol-based referencing, we instruct an LLM to act as an information fusion tool for generating a coherent social story $\bm{S}$ in a unified paragraph. The social story tells all the information needed about the visual scene for text-based reasoning, which is highly readable and understandable by humans and LLMs with clear symbol references and information consolidation. An example of extracted perceptual information with symbol associations and the generated social story is depicted in Figure~\ref{fig:dense_caption}.

\subsection{Reasoning with Large Language Models}

After obtaining the mapping from image to social story: $\bm{I} \rightarrow \bm{S}$, we feed both $\bm{S}$ and bounding box queries $(\bm{b}_i, \bm{b}_j)$, converted to textual queries $\bm{q}$ with referencing symbols $P_i, P_j$, into LLMs to obtain interpretable answers $\bm{a}$. This is to let LLMs output the map from $(\bm{S}, \bm{q})$ to $\bm{a}$, which we do by prompting. Since LLM performance is highly sensitive to prompt variations~\cite{white2023prompt,white2023chatgpt}, we design our social relation reasoning prompt with four segments, which we name {\namep}.

\noindent\textbf{System.}
This is the system prompt provided by many LLMs to steer their behavior. We utilize it to explicitly define several core rules for our task of social reasoning. We denote it as the $\bm{o}$ segment.

\noindent\textbf{Expectation.}
This is the instruction that we give to the model to set expectations of the anticipated outcomes. This helps avoid vague or unexpected outputs. To do so, we construct a role assignment and task description prompt, denoted as $\bm{r}$, where we explicitly assign the role of a social relation expert to the LLM and provide a detailed elaboration of the task's input and output.

\noindent\textbf{Context.}
This provides sufficient contextual information to help the LLMs understand the background of the problem. As a classification task, we provide specific definitions for each social relationship category, resulting in the prompt segment denoted as $\bm{c}$.

\noindent\textbf{Guidance.}
This offers an exemplar to show the LLMs how to respond to a query based on a social story. In-context learning has been proven as an effective means to expand the capabilities of LLMs~\cite{brown2020language,min2022rethinking,rubin2022learning}. We manually construct an in-context example prompt, denoted as $\bm{e} = (\bm{S}_0, \bm{q}_0, \bm{a}_0)$, to better guide LLMs in performing social relationship reasoning in the desired format. Here we also guide the model to generate possible explanations for its prediction. While using more in-context examples may potentially further enhance performance, this is beyond the scope of the paper and is left as future work.

The final {\namep} consists of $(\bm{o}, \bm{r}, \bm{c}, \bm{e})$, and is concatenated with a testing story-query pair $(\bm{S}, \bm{q})$ at the end for model predictions.
Figure~\ref{fig:framework} shows the structured excerpts of {\namep}, and we put the full prompt into the Appendix.
Note that we do not use any training samples provided by a dataset and only employ the foundation models. Consequently, {\name} is capable of zero-shot social relation reasoning, while maintaining its interpretability and generalizability.

\section{Greedy Segment Prompt Optimization}

Although we have devised well-structured {\namep} for social relation reasoning, experiments reveal that different ways of prompt rephrasing and demonstration example variations can significantly impact the LLM reasoning performance.
Manually searching for the optimal prompt is time-consuming and labor-intensive, thus automatic prompt tuning is desired.
Nevertheless, unlike the prompt optimization methods~\cite{zhou2022large,pryzant-etal-2023-automatic} typically employed in NLP, automatic prompt tuning for {\namep} faces two unique challenges: \textit{free-form target} and \textit{long prompt optimization}. As we convert a visual classification task into a generative task for LLMs, the model's output space transitions from discrete numerical representations of one-hot labels to unconstrained textual forms. Defining free-form text objectives for {\namep} optimization is not well-explored. 
Meanwhile, as the social story $\bm{S}$ is a comprehensive description of the image such as in Figure~\ref{fig:dense_caption}, and task and full label set definitions could be lengthy, our {\namep} tends to be very long. This poses additional challenges for automatic prompt tuning methods.
To address these issues, we propose a segment-based optimization algorithm, named Greedy Segment Prompt Optimization (GSPO).

\begin{algorithm}[t]
\caption{Greedy Segment Prompt Optimization}
\label{alg:gsd}
\begin{algorithmic}
\Require Initial segments $\bm{w}_{1:M}$, training dataset $\mathcal{T}$, iteration number $N$
    \State Build the candidate set $\mathcal{W}_m$ for each segment $\bm{w}_m$
    \Loop{ $N$ times 
    }
        \State Randomly sample a batch of data $\mathcal{D}$ from $\mathcal{T}$
        \For{$m = 1,\ldots,M $}
            \State $\mathcal{U}_m := \mbox{Top-}k(- \sum_{\bm{z} \in \mathcal{D} } \nabla_{h_{w_m}} \mathcal{L}(\bm{w}_{1:M};\bm{z}))$ \\ \Comment{Compute top-$k$ promising segment substitutions}
        \EndFor
        \For{$b = 0, 1,\ldots, K*M -1 $}
            \State $\tilde{\bm{w}}_{1:M}^{(b)} := \bm{w}_{1:M}$
            \Comment{Initialization}
            \State $\tilde{w}^{(b)}_{i} := \mathcal{U}_i( \lfloor b / M \rfloor)$, where $i = ( b \bmod M) + 1$ \\ \Comment{Select one replacement segment}
        \EndFor
        \State $\bm{w}_{1:M} := \tilde{\bm{w}}^{(b^\star)}_{1:M}$, where $b^\star = \argmin_b \sum_{\bm{z} \in \mathcal{D} } \mathcal{L}(\tilde{\bm{w}}^{(b)}_{1:M},\bm{z})$ \Comment{Compute best replacement}
    \EndLoop
\Ensure Optimized segments $\bm{w}_{1:M}$
\end{algorithmic}
\end{algorithm}

\noindent\textbf{Tuning Objective.} 
To automate prompt searching, the first step is to define the optimization objective. Ideally, we aim to find the optimal prompt $\{\bm{o}^*, \bm{r}^*, \bm{c}^*, \bm{e}^*\}$  that maximize the probability of LLMs generating the correct answer $\bm{a}$ for any given sample $\bm{z} =(\bm{S}, \bm{q})$. Let's first review the training paradigm commonly used for autoregressive language models~\cite{radford2018improving,radford2019language,brown2020language}, which essentially employ the next token prediction task, \emph{i.e.}, learning $p(w_{n+1} | w_{1:n})$, where token $w_{n+1} \in \mathcal{V}$, and $\mathcal{V}$ represents the token vocabulary. 
Unlike typical classification tasks where only a one-hot formatted category is predicted, our answers are free-form text, consisting of a sequence of numerous tokens.  Constructing the ground truth with free-form text for each sample is challenging. This paper proposes instructing LLMs to begin their response with the predicted class category following a pre-defined template. Formally, we assume that the ground truth answer $\bm{a}$ for sample $\bm{z}$ takes the following form: $\bm{a} = [\bm{a}^0,\bm{a}^1,\bm{a}^2, ...]$, where $\bm{a}^0$ denotes the first sentence of $\bm{a}$, $\bm{a}^1$ is the second sentence, and so forth. We specify $\bm{a}^0$ to have the following fixed format: $\bm{a}^0 = $  ``\emph{The final answer is} $str(\bm{y})$", where $str(\bm{y})$ represents the string representation of class label $\bm{y}$. Then we can define the objective:
\begin{equation}
    \mathcal{L} (\bm{o}, \bm{r}, \bm{c}, \bm{e}; \bm{z},\bm{y}) = - \mathbb{E}_{(\bm{z}, \bm{a}^0)} \left[ \log p(\bm{a}^0 | \bm{o}, \bm{r}, \bm{c}, \bm{e}; \bm{z}) \right],
    \label{eq:objective}
\end{equation}
where the expectation is taken from a collection of training examples, and the probabilities are computed from LLM's next token prediction distributions. 
Note here the LLM is frozen, and we seek to find the optimal prompt to minimize the above loss.
In practice, we employ the same template in our in-context example, making it easy for LLMs to follow a consistent output format. This ensures that the loss primarily stems from LLMs' predictions of tokenized category names rather than category-agnostic sentence formatting. Note that we only construct and supervise the first sentence of the ground truth answer, while the model is free to generate its explanation in the following sentences.

\noindent\textbf{Long Prompt Optimization.}
We optimize over discrete prompt tokens, constrained to a vocabulary $\mathcal{V}$ for each token position associated with the LLM.
While some discrete prompt optimization algorithms~\cite{zou2023universal,shin2020autoprompt,zou2023universal} have been proposed in the NLP field, they typically operate on a limited number of tokens. In contrast, as a visual reasoning task, we require long prompts to adequately convey the dense information and provide detailed context.
In fact, the number of tokens in our {\namep} may well exceed 2K, and conduct token-level optimization results in a search space of $2000^{|\mathcal{V}|}$, which is beyond the capacities of current optimization methods as $|\mathcal{V}| = 32,000$ for many LLMs~\cite{touvron2023llama,touvron2023llama2}.
We propose to perform segment-level optimization as a surrogate.
Formally, suppose the prompt is $\bm{w}$ with $M$ segments, denoted as $\bm{w}_{1:M}$.
In our case we can have $M=4$ and directly map the segments to $\bm{o}, \bm{r}, \bm{c}, \bm{e}$, respectively. We propose a candidate set $\mathcal{W}_m$ consisting of alternative prompts for each segment, which we use ChatGPT to generate followed by light manual revisions, and the algorithm searches over the combination of different candidates.
For the demonstration example segment $\bm{e}$, we also manually select samples from an existing 
training set as candidates.

More specifically, inspired by AutoPrompt~\cite{shin2020autoprompt}, our optimization algorithm considers all possible single-segment substitutions, thereby selecting the segment candidate that minimizes the loss over a batch of training samples.
We replace one segment at a time in a greedy manner.
In practice, instead of evaluating all possible candidates, we further reduce the search space by calculating the gradients of the one-hot segment indicators for each segment and selecting the top $K$ most promising candidates for that segment. The gradient is computed as: $ \nabla_{h_{w_m}} \mathcal{L}(\bm{w}_{1:M}) \in \mathbb{R}^{|\mathcal{W}_m|}$, where $h_{w_m}$ represents the one-hot representation of selecting $\bm{w}_m$ from the set $\mathcal{W}_m$. Then the top $K$ promising substitutions with the largest negative gradient are chosen for evaluation. We repeat this process to acquire $K$ candidates for each segment, and we only replace one segment at a time to obtain $K * M$ new prompts. Then the one with the smallest loss over a batch of training samples is chosen. We iterate this process $N$ times to find the best-performing prompt.
The entire search process is shown in Algorithm \ref{alg:gsd}.

\begin{table}[t]
\parbox{.45\linewidth}{
\centering
  \renewcommand\tabcolsep{5pt}
    \caption{The comparison results on the PIPA dataset. ZS stands for Zero-Shot.}
  \label{table:sotaPIPA}
  \centering
  \begin{tabular}{lcc}
    \toprule
    Methods &  ZS & Acc (\%) \\
    \midrule
    All attributes + SVM \cite{sun2017domain} & \textcolor{ForestGreen}{\XSolidBrush} & 57.2 \\
     Pair CNN \cite{li2017dual}& \textcolor{ForestGreen}{\XSolidBrush}& 58.0 \\
     Dual-Glance  \cite{li2017dual} & \textcolor{ForestGreen}{\XSolidBrush}& 59.6 \\
     SRG-GN \cite{goel2019end}  & \textcolor{ForestGreen}{\XSolidBrush}& 53.6 \\
     GRM \cite{wang2018deep}  & \textcolor{ForestGreen}{\XSolidBrush}& 62.3 \\
     MGR \cite{zhang2019multi} & \textcolor{ForestGreen}{\XSolidBrush}&  64.4 \\
     GR$^2$N~\cite{li2020graph} &\textcolor{ForestGreen}{\XSolidBrush} &  64.3 \\
     TRGAT~\cite{guo2023social} &\textcolor{ForestGreen}{\XSolidBrush} & 65.3 \\
     \midrule
     {\name} (w/ GPT-3.5) & \textcolor{LavaRed}{\CheckmarkBold} & 64.1 \\
     {\name} (w/ Vicuna-13B) & \textcolor{LavaRed}{\CheckmarkBold} &  \textbf{66.7} \\
    \bottomrule
  \end{tabular}
}
\hfill
\parbox{.5\linewidth}{
\centering
    \renewcommand\tabcolsep{5pt}
  \caption{Ablations on components of {\name} with Vicuna-7B. The results are obtained on the PIPA dataset with a zero-shot setting.} 
  \label{table:ablation}
  \centering
    \begin{tabular}{lc}
    \toprule 
     Methods & Acc (\%) \\
    \midrule 
     {\name}   & \textbf{61.58} \\
    \midrule 
    - Dense Captions & 52.63 \\
    - Task-oriented Captions & 59.89 \\
    - Symbol $\rightarrow$ Object Coordinate & 57.68 \\
    - Symbol $\rightarrow$ Object Caption & 59.83 \\
    - Social Story & 45.31 \\
    \midrule
    - {\namep} Segment \{System\} & 60.23\\
    - {\namep} Segment \{Expectation\} & 59.19 \\
    - {\namep} Segment \{Context\} & 61.18 \\
    - {\namep} Segment \{Guidance\} & 43.56 \\
    \bottomrule
  \end{tabular}
}
% \vspace{-16pt}
\end{table}

\section{Experiments}
\label{sec:experiment}

\subsection{Settings}

\noindent\textbf{Data and Evaluation.}
We adopt two widely-used benchmarks for social relation reasoning: PIPA~\cite{sun2017domain} and PISC~\cite{li2017dual}.
The PIPA dataset categorizes 16 types of social relationships, including family bonds (like parent-child, grandparent-grandchild), personal connections (friends, loves/spouses), educational and professional interactions (teacher-student, leader-subordinate), and group associations (band, sports team, colleagues).
The PISC dataset categorizes social relationships into six types: commercial, couple, family, friends, professional, and no-relation.
We follow the standard train/val/test split for both datasets and report the classification accuracy on the test set.
Note that the training set is not used for our zero-shot results, but is used for in-context exemplar proposals for our prompt optimization algorithm.
For both datasets, we measure classification accuracy as our evaluation metric.

\noindent\textbf{Implementation Details.}
We use two VFM models for visual information extraction -- the SAM~\cite{kirillov2023segment} model for object segmentation, followed by BLIP-2~\cite{li2022blip} for dense caption generation. For the social story generation, we employ the GPT-3.5~\cite{white2023chatgpt} Turbo model that has empowered ChatGPT. We set the temperature to 0 for greedy decoding to bolster the result's reproducibility. Other generation parameters are otherwise set as default.
For subsequent reasoning of social relations based on generated stories, we experiment with both GPT-3.5 and open-source LLMs, including Vicuna-7B/13B~\cite{vicuna2023} and Llama2-7B/13B~\cite{touvron2023llama2}. All the decoding temperature is set as 0, and we set the maximum context length to 4096 for Vicuna and Llama2 to accommodate our long prompt.
For GSPO, we curate $M=15$ candidates for each of the four segments within the complete prompt and set $K=3$ for candidate selection for $N=500$ iterations. One A100 GPU is used for all experiments.

\subsection{Zero-shot Social Relation Recognition with {\name}}

\noindent\textbf{Main Results.}
We compare {\name}, using either GPT-3.5 or Vicuna-13B, with previous fully supervised methods and present our results in Table~\ref{table:sotaPIPA} and Table~\ref{table:sotaPISC}. Here our method does not undergo the prompt tuning optimization, performing relation reasoning in a zero-shot fashion without utilizing any training examples.
On both datasets, Vicuna-13B performs better than GPT-3.5 with our framework. In particular, on PIPA benchmark shown in Table~\ref{table:sotaPIPA}, {\name} achieves the best accuracy compared with all prior supervised approaches, leading the previous state-of-the-art model TRGAT~\cite{guo2023social} by 1.4\%.
The results on the PISC benchmark are shown in Table~\ref{table:sotaPISC}. 
Most previous methods used mAP (mean Average Precision) as the metric on the PISC dataset, whereas we opted not to employ this metric due to the disparity between our predictions. Unlike previous methods that output per-class confidence scores, our prediction is the textual outputs from LLMs. Therefore, we still adopt the accuracy metric on the PISC dataset. To report the accuracy performance of other methods, we chose the state-of-the-art methods with publicly available code for reproduction and compared their performance.
Table~\ref{table:sotaPISC} shows that our method attains comparable results to the state-of-the-art GR$^2$N model, despite not being trained with any data.

\noindent\textbf{Comparison with End-to-End VLMs.}
Our approach breaks down the social relation reasoning into different phases involving perception tasks with VFMs and reasoning with LLMs, bridged by a coherent textual social story. However, recent advancements in multimodal foundation models (VLMs) provide a straightforward way of reasoning about visual contents, which is simply asking questions about the image to a vision-language model that can respond with an answer directly. We compare {\name} with three state-of-the-art end-to-end vision-language foundation models by directly inquiring about social relationships in the image, including BLIP-2~\cite{li2022blip}, LLaVA~\cite{liu2023visual-llava}, and GPT-4V~\cite{white2023chatgpt}, with results shown in Table~\ref{table:vlm}. We see that the method of querying vision-language foundation models, albeit simple, is still lagging behind our approach of {\name} with principled designs and modularized VFMs and LLMs. Our well-designed {\name} even outperforms the high-performing GPT-4V by 7.03\% in accuracy. These results justify the design principles of our framework with comprehensive perception extraction and coherent language reasoning.

\begin{table}[t]
\parbox{.58\linewidth}{
\centering
  \renewcommand\tabcolsep{8pt}
    \caption{The comparison results on the PISC dataset.
    Previous methods are replicated with open-source code to report the accuracy metric. ZS means Zero-Shot.}
  \label{table:sotaPISC}
  \centering
  \begin{tabular}{lcc}
    \toprule
    Methods & ZS &  Acc (\%) \\
    \midrule
     Pair CNN \cite{li2017dual}& \textcolor{ForestGreen}{\XSolidBrush}&  46.30 \\
     GRM \cite{wang2018deep} & \textcolor{ForestGreen}{\XSolidBrush} & 64.18 \\
     GR$^2$N~\cite{li2020graph}& \textcolor{ForestGreen}{\XSolidBrush} & 64.70 \\
     \midrule
     {\name} (w/ GPT-3.5)& \textcolor{LavaRed}{\CheckmarkBold} & 53.43 \\
     {\name} (w/ Vicuna-13B) & \textcolor{LavaRed}{\CheckmarkBold} & \textbf{65.12} \\
    \bottomrule
  \end{tabular}
}
\hfill
\parbox{.37\linewidth}{
\centering
    \renewcommand\tabcolsep{8pt}
  \caption{Comparison with existing Vision-Language Models on the PIPA dataset, with {\name} using Vicuna-13B model.}
  \label{table:vlm}
  \centering
  \begin{tabular}{lc}
    \toprule
    Methods & Acc (\%) \\
    \midrule
     BLIP-2~\cite{li2022blip} & 35.84 \\
     LLaVA~\cite{liu2023visual-llava} & 45.12 \\
     GPT-4V~\cite{white2023chatgpt} & 59.67 \\
     \midrule
     {\name} & \textbf{66.70} \\
    \bottomrule
  \end{tabular}
}
% \vspace{-16pt}
\end{table}

\noindent\textbf{Ablation Study.}
We conduct a series of ablation studies to assess the efficacy of various components at different stages of {\name}. Table~\ref{table:ablation} shows the results with Vicuna-7B on the PIPA dataset.
The first part of ablation focuses on the social story generation pipeline.
As we use SAM to segment the image for visual perception, removing SAM would disable fine-grained object descriptions (dense captions) in the social story, resulting in an accuracy drop of more than 8\%.
If we do not acquire the task-oriented captions,
there is a performance drop of 1.69\%.
Next, a crucial component of the social story generation in {\name} is the utilization of symbols ($P$ for people and $O$ for others) for effective referral of objects. If we do not use the symbols, but instead replace the object referral with either the direct coordinate or the object-specific caption from BLIP-2 in both the social story and the question, we see the performance drops by 3.90\% and 1.75\%, respectively.
Finally, we fuse the multi-aspect visual information into a cohesive social story.
If we bypass the fusion and directly utilize the visual annotations from VLMs, we can see there is a significant performance drop of 16.27\%. This indicates that a good textual description of comprehensive visual information is necessary to connect LLMs to reason about social scenes presented in images.

We also ablate the {\namep} segments in our LLM reasoning phase. We do this by removing each of the segments from the full prompt one at a time, and results are presented in the bottom half of Table~\ref{table:ablation}. We can see that guidance segmentation, which contains a manually constructed demonstration example of how to reason about social relations based on our social story, has the most influence on the model performance. Without it, the accuracy drops by 18.02\%. The system prompt and expectation segment contributes to the final performance by approximately 1.35\% and 2.39\%, respectively, and the context segment defining social relationship categories has a lesser contribution with a 0.4\% accuracy difference. This is perhaps because the LLMs already have substantial knowledge of common social relationships.

\begin{table}[t]
  \renewcommand\tabcolsep{7pt}
      \caption{Prompt tuning results (accuracy in \%) with GSPO. 
    }
  \label{table:promptoptipipa+pica}
  \centering
  \begin{tabular}{lccc|ccc}
    \toprule
    \multirow{2}{*}{Model}& \multicolumn{3}{c|}{PIPA} & \multicolumn{3}{c}{PISC}  \\
    \cmidrule{2-4} \cmidrule{5-7}
     & {\name} & + GSPO  & $\Delta$ & {\name} & + GSPO  & $\Delta$ \\
    \midrule
     Vicuna-7B  & 61.58 & 62.99 & \textcolor{OliveGreen}{+1.41} & 45.13 & 49.79 & \textcolor{OliveGreen}{+4.66} \\
     Vicuna-13B & \textbf{66.70} & \textbf{69.23} & \textcolor{OliveGreen}{+2.53} & \textbf{65.12} & \textbf{66.19} & \textcolor{OliveGreen}{+1.07} \\
     Llama2-7B & 31.91 & 34.07  & \textcolor{OliveGreen}{+2.16} & 36.71 & 38.04 & \textcolor{OliveGreen}{+1.33} \\
     Llama2-13B & 37.86 & 41.27 & \textcolor{OliveGreen}{+3.41}  & 42.74 & 48.39 & \textcolor{OliveGreen}{+5.65}\\
    \bottomrule
  \end{tabular}
  % \vspace{-12pt}
\end{table}

\subsection{Long Prompt Optimization with GSPO}

As {\name} utilizes fixed prompt segments to instruct LLMs for social relation reasoning based on social stories, it might not be optimal with the static prompt design. Our GSPO further tunes the long prompt on the segment level for automatic performance improvements.
Table~\ref{table:promptoptipipa+pica} presents the results when applying GSPO on {\name} with various LLMs for reasoning, compared with the baseline zero-shot performance. Overall our segment-level prompt tuning with GSPO helps with the classification of all model variants. On PIPA the performance boost is about 2.38\% on average, and on PISC it achieves a better gain with about 3.18\% on average. These show the efficacy of the proposed GSPO algorithm to efficiently enhance prompt effectiveness.
Out of the model variations, Vicuna-13B consistently outperforms other LLMs under our setup.
The flexibility of {\name} in connecting with different reasoning models makes it more easily benefit from the latest advancements of LLMs without any heavy adaptation.

\subsection{Qualitative Analysis}

\begin{figure}[t]
\centering
\includegraphics[width=1.0\linewidth]{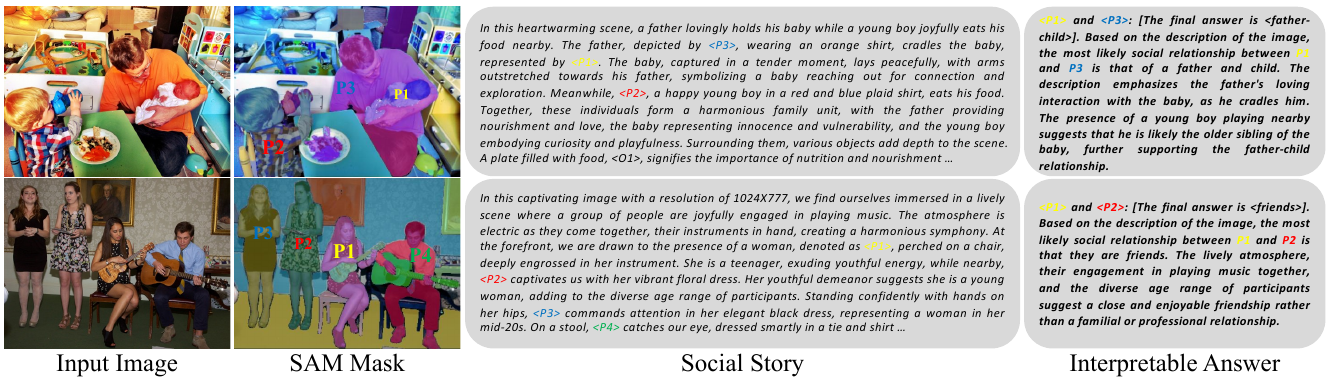}
\caption{Visualization results of interpretability. We show the {\name} perception and reasoning process. 
We see that our model predicts correct social relationships with plausible explanations.}
\label{fig:vis_interpretability}
\vspace{-14pt}
\end{figure}

\noindent\textbf{Reasoning Process and Interpretability.}
We illustrate the perception and reasoning process of {\name} as well as the final results in Figure~\ref{fig:vis_interpretability}. The people objects are fully segmented from VFMs and associated with symbols, which are then utilized to generate a coherent social story with clear references. By using LLMs for the reasoning on top of textual stories, {\name} not only outputs the correct social relations between different objects in the image but also provides plausible explanations behind the reasoning process.

\begin{figure}[t]
\centering
\includegraphics[width=1.0\linewidth]{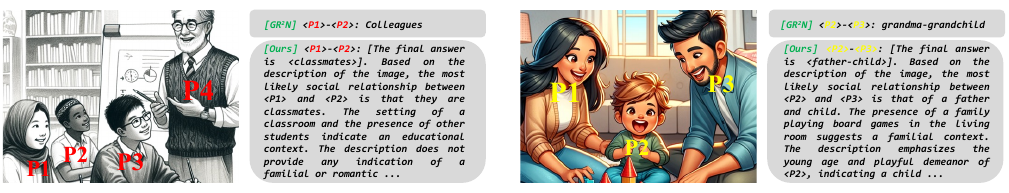}
\caption{Results when applying {\name} to sketch and cartoon images. 
The images are generated by GPT-4V. Our method generalizes well on these novel image styles.}
% \vspace{-2mm}
\label{fig:vis_generality}
% \vspace{-18pt}
\end{figure}

\noindent\textbf{Generalization on Different Image Styles.}
Previous supervised models on social relation recognition heavily rely on annotated images and relations in a specific domain.
As a result, these models cannot generalize to unseen image types well. In contrast, our method does not have the limitation of being domain-specific. We apply {\name} to novel sketch and cartoon images with various social relations generated by GPT-4V, with results shown in Figure~\ref{fig:vis_generality}. As shown in the first example, 
the previous state-of-the-art model GR$^2$N~\cite{li2020graph} fails to generalize as it predicts the relation between $P_1$ and $P_2$ as colleagues, but {\name} correctly recognizes the classmate relation based on the social scene with detailed explanation.

\section{Conclusion}
\label{sec:conclusion}

\noindent \textbf{Conclusion.} In this paper, we present {\name}, a modular framework with foundation models for social relation reasoning,
which attains competitive zero-shot results while also providing interpretable explanations.
Furthermore, we propose the GSPO 
for automatic prompt tuning, which further improves the performance.
Our approach opens new avenues for exploring the synergy between vision and language models in high-level cognitive tasks and 
offers a promising direction for future advancements in the field of social relation recognition.

\textbf{Limitations and broader impacts.} Due to the modular nature of our approach,  the performance of our method is constrained by the performance of the foundation models. If the segmentation model fails, or if the BLIP-2 model generates incorrect captions, or if the reasoning by LLMs is flawed, then our method is also prone to errors. 
Our method transforms visual problems into language-based reasoning, which could improve accessibility for visually impaired individuals.
Meanwhile, our method also inherits biases from the foundation models, thus further research is needed to address them.
Automatic classification of social relationships may lead to unintended negative consequences. To mitigate these risks, we can implement strategies such as fairness and bias checks, as well as promote transparent and responsible use of our technology.

\section*{Acknowledgment}
This research is supported in part by the NIH grant R01HD104969, NIH grant 1U01CA284207, and NSF award IIS-2239688.

{\small
\bibliographystyle{unsrt}
\bibliography{egbib}
}
\clearpage

%%%%%%%%%%%%%%%%%%%%%%%%%%%%%%%%%%%%%%%%%%%%%%%%%%%%%%%%%%%%

\appendix

% % \section{Appendix}
% \section{Algorithm}

% The proposed greedy segment prompt optimization process is shown in Algorithm \ref{alg:gsd}.

% \begin{algorithm}[h]
% \caption{Greedy Segment Prompt Optimization}
% \label{alg:gsd}
% \begin{algorithmic}
% \Require Initial segments $\bm{w}_{1:M}$, training dataset $\mathcal{T}$, iteration number $N$
%     \State Build the candidate set $\mathcal{W}_m$ for each segment $\bm{w}_m$
%     \Loop{ $N$ times 
%     % \jiawei{is I here M? symbols are a bit confusing} \wanhua{No, it's iterations, I add it in the InPut now}
%     }
%         \State Randomly sample a batch of data $\mathcal{D}$ from $\mathcal{T}$
%         \For{$m = 1,\ldots,M $}
%             \State $\mathcal{U}_m := \mbox{Top-}k(- \sum_{\bm{z} \in \mathcal{D} } \nabla_{h_{w_m}} \mathcal{L}(\bm{w}_{1:M};\bm{z}))$ \\ \Comment{Compute top-$k$ promising segment substitutions}
%         \EndFor
%         \For{$b = 0, 1,\ldots, K*M -1 $}
%             \State $\tilde{\bm{w}}_{1:M}^{(b)} := \bm{w}_{1:M}$
%             \Comment{Initialization}
%             \State $\tilde{w}^{(b)}_{i} := \mathcal{U}_i( \lfloor b / M \rfloor)$, where $i = ( b \bmod M) + 1$ \\ \Comment{Select one replacement segment}
%         \EndFor
%         \State $\bm{w}_{1:M} := \tilde{\bm{w}}^{(b^\star)}_{1:M}$, where $b^\star = \argmin_b \sum_{\bm{z} \in \mathcal{D} } \mathcal{L}(\tilde{\bm{w}}^{(b)}_{1:M},\bm{z})$ \Comment{Compute best replacement}
%     \EndLoop
% \Ensure Optimized segments $\bm{w}_{1:M}$
% \end{algorithmic}
% \end{algorithm}

\section{More Implementation Details}

\begin{figure*}[h]
\centering
\includegraphics[width=1.0\linewidth]{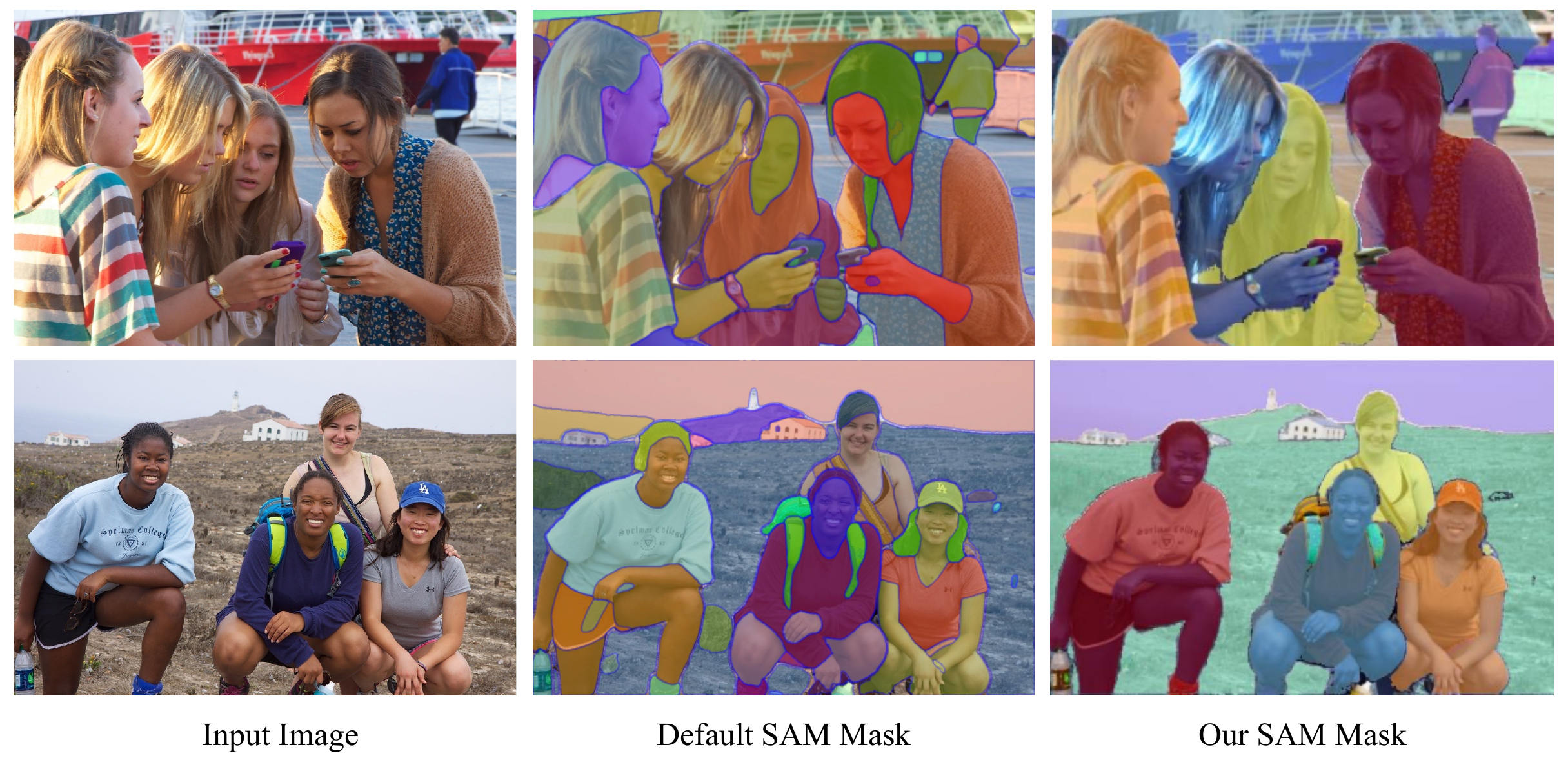}
% \vspace{-1in}
\caption{The comparisons of the default SAM masks and our SAM masks.}
\label{fig:sammask}
\end{figure*}

In this paper, we employ SAM to automatically segment an image into multiple object masks, which we then use to generate dense captions. However, a challenge arises with SAM's default ``segment everything" setting, as it tends to produce over-segmented and fine-grained masks. For instance, a person may be segmented into multiple fragments, including hair, face, hand, arm, and so on. Two examples illustrating this issue are presented in Figure \ref{fig:sammask}. Creating meaningful captions for these subpart-level regions proves to be challenging and often leads to a loss of overall object perception. This is due to the fact that SAM generates three masks for each point prompt, corresponding to three semantic levels: whole, part, and sub-part. 
To address this issue, we adopted a two-stage SAM forward scheme. Initially, we employed SAM's default ``segment anything" approach to obtain segmented masks, then retained the center points of each mask as point prompts for the second SAM forward pass. This ensures that as much as possible, objects in the image are not missed in the second SAM segmentation stage. For the second SAM segmentation stage, the points obtained from the first stage are used as point prompts, considering only the highest semantic level among SAM's three semantic levels. This approach minimizes over-segmentation and allows our method to focus on semantic at the object level. Subsequently, we apply NMS, threshold filtering, and post-processing to obtain high-quality object-level masks following SAM's methodology~\cite{kirillov2023segment}. The resulting object masks for our method are displayed in Figure \ref{fig:sammask}.

\begin{figure*}[ht!]
    \centering
    \includegraphics[width=1.0\textwidth]{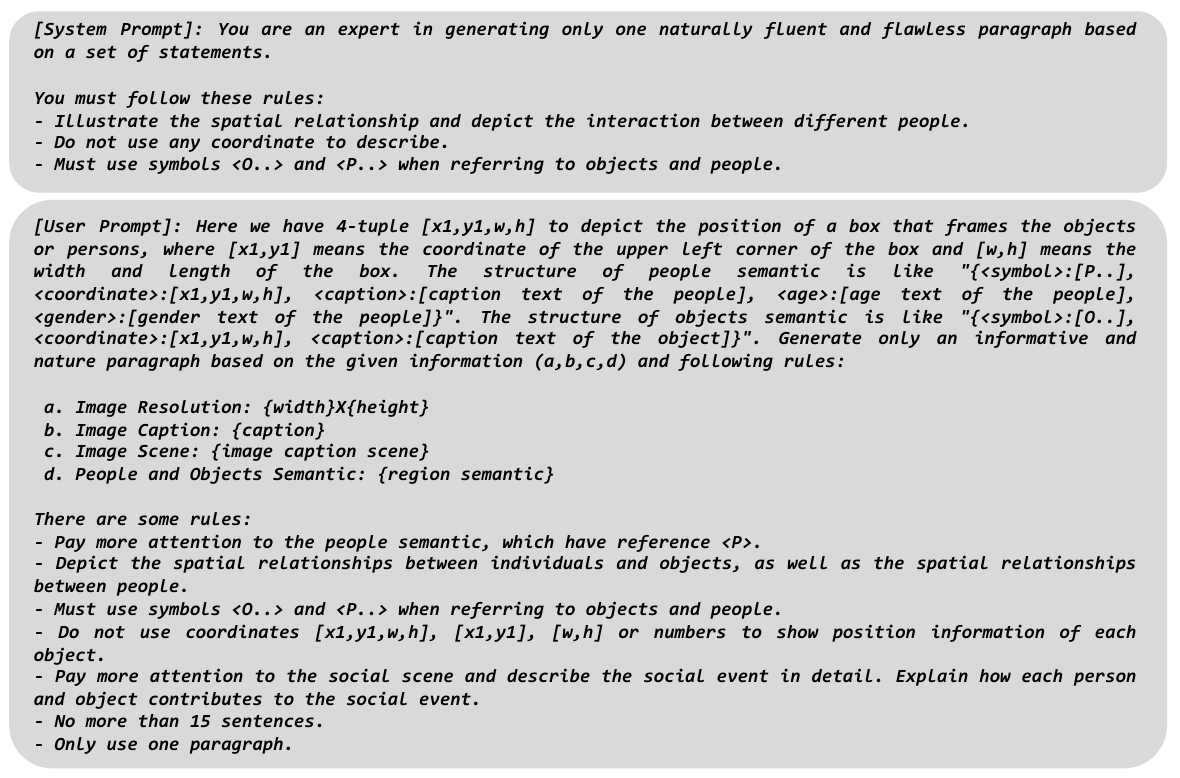}
    \caption{ The prompt used for social story generation. GPT-3.5 Turbo model is used for caption fusion. The system prompt lists some key rules and the user prompt details the task definition.}  
    \label{fig:social}
\end{figure*}

\section{Prompts}
\noindent\textbf{Social Story Generation.} We carefully designed the prompt to guide the LLMs in generating coherent and easily understandable social stories based on dense captions. The system prompt and user prompt are depicted in Figure \ref{fig:social}. To ensure symbol-based referencing, we explicitly instruct LLMs not to rely on coordinates but instead to use symbols for reference. Additionally, we require the generated paragraphs to focus on social contexts.

\begin{figure*}[t]
    \centering
    \includegraphics[width=1.0\textwidth]{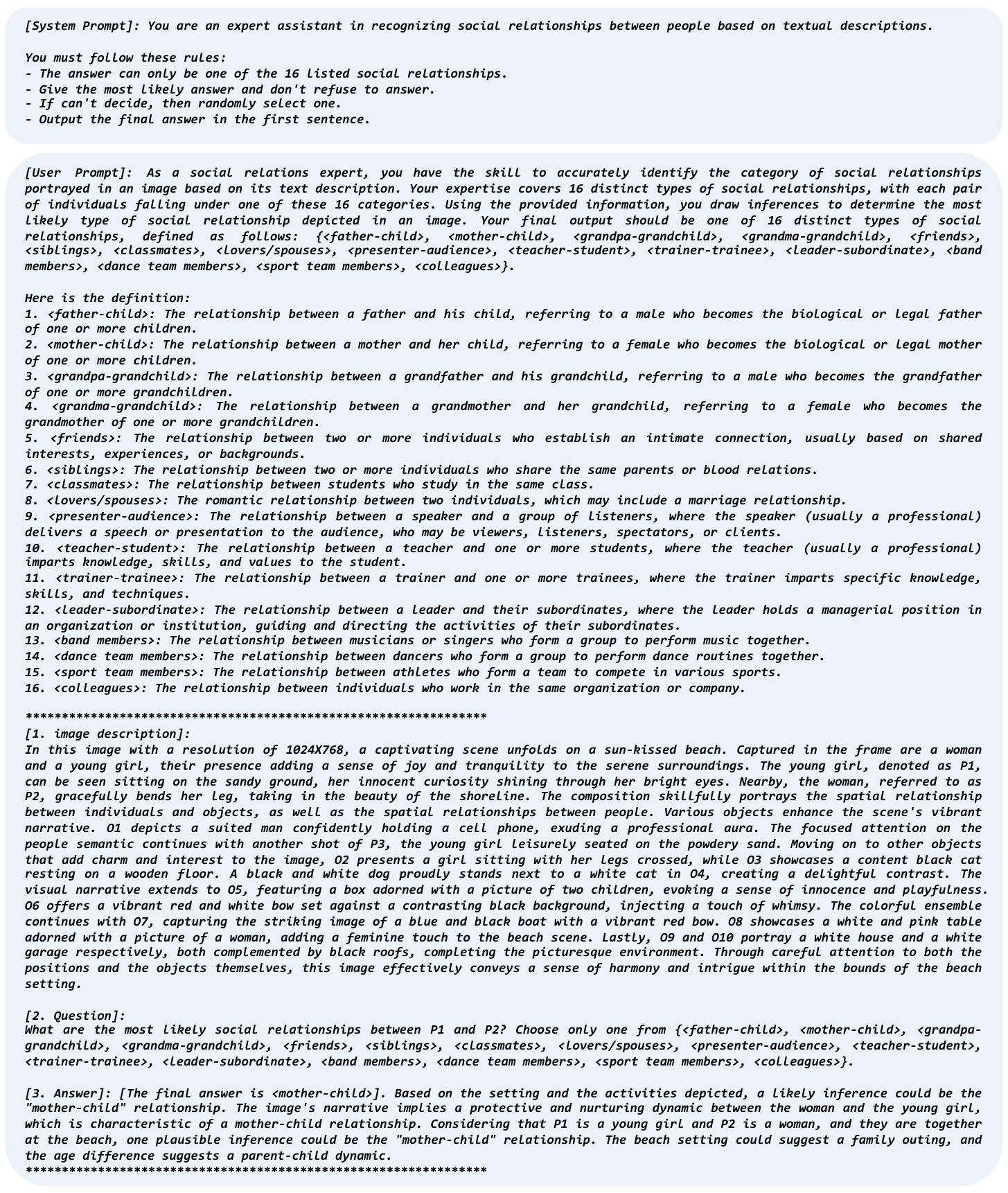}
    \caption{ The {\namep} on the PIPA dataset. 
    }  \label{fig:pipainitial}
\end{figure*}

\noindent\textbf{{\namep} on the PIPA dataset.} The PIPA dataset comprises 16 social relationship categories, including father-child, mother-child, grandpa-grandchild, grandma-grandchild, friends, siblings, classmates, loves/spouses, presenter-audience, teacher-student, trainer-trainee, leader-subordinate, band members, dance team members, sport team members, and colleagues. Figure \ref{fig:pipainitial} illustrates the prompt we utilized for the PIPA dataset in the zero-shot setting. We provided a detailed explanation for each category within the prompt. Furthermore, the {\namep} includes manually constructed in-context examples. 

\begin{figure*}[t]
    \centering
    \includegraphics[width=1.0\textwidth]{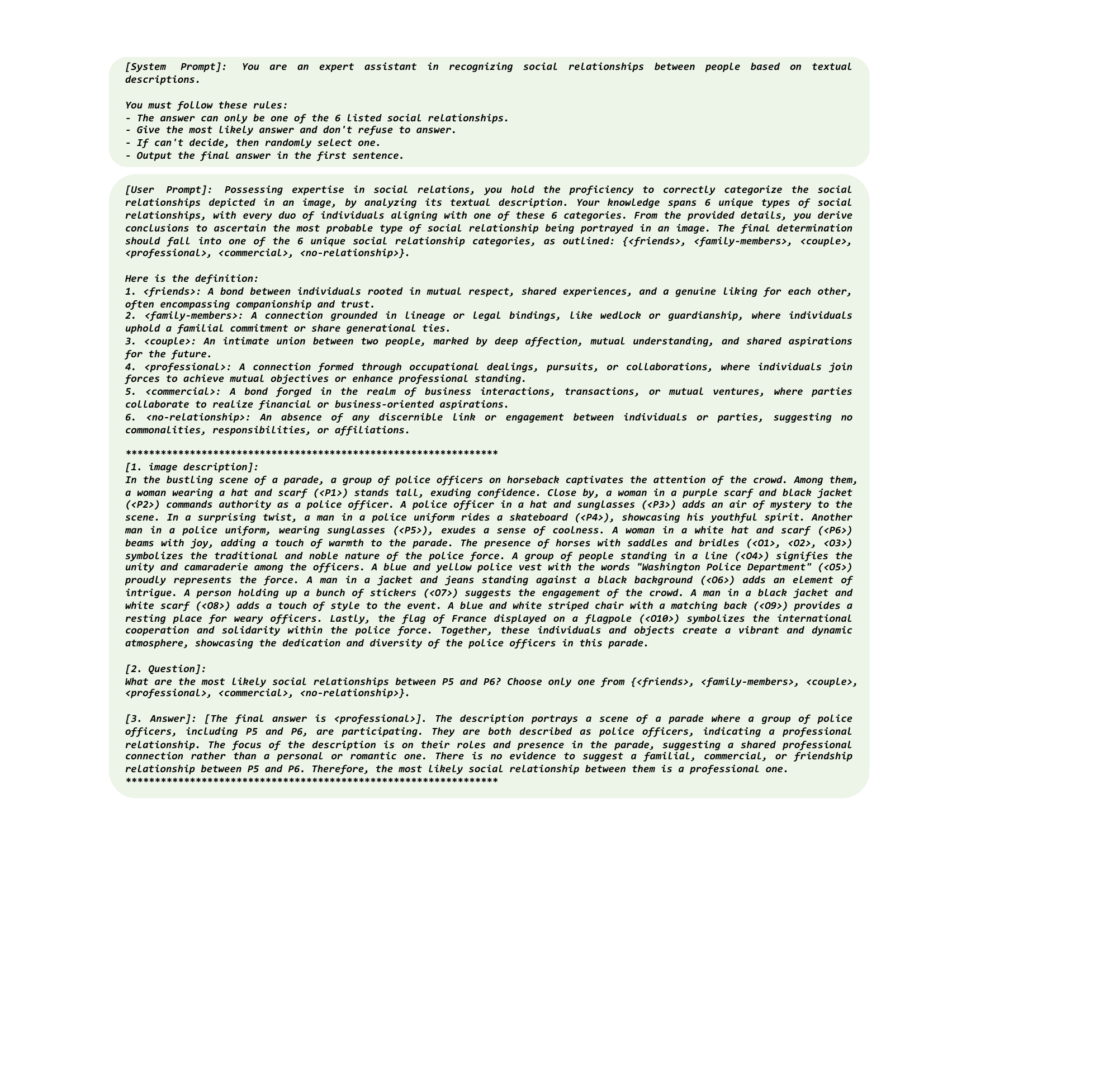}
    \caption{ The {\namep} on the PISC dataset. 
    }  \label{fig:piscinitial}
\end{figure*}

\noindent\textbf{{\namep} on the PISC dataset.} Figure \ref{fig:piscinitial} illustrates the {\namep} utilized in the PISC dataset, specifically in the zero-shot setting. The PISC dataset comprises 6 social relation categories: commercial, couple, family, friends, professional, and no-relation. We have also included the definitions of these six social relation categories within the prompt. 

\begin{figure*}[t]
    \centering
    \includegraphics[width=1.0\textwidth]{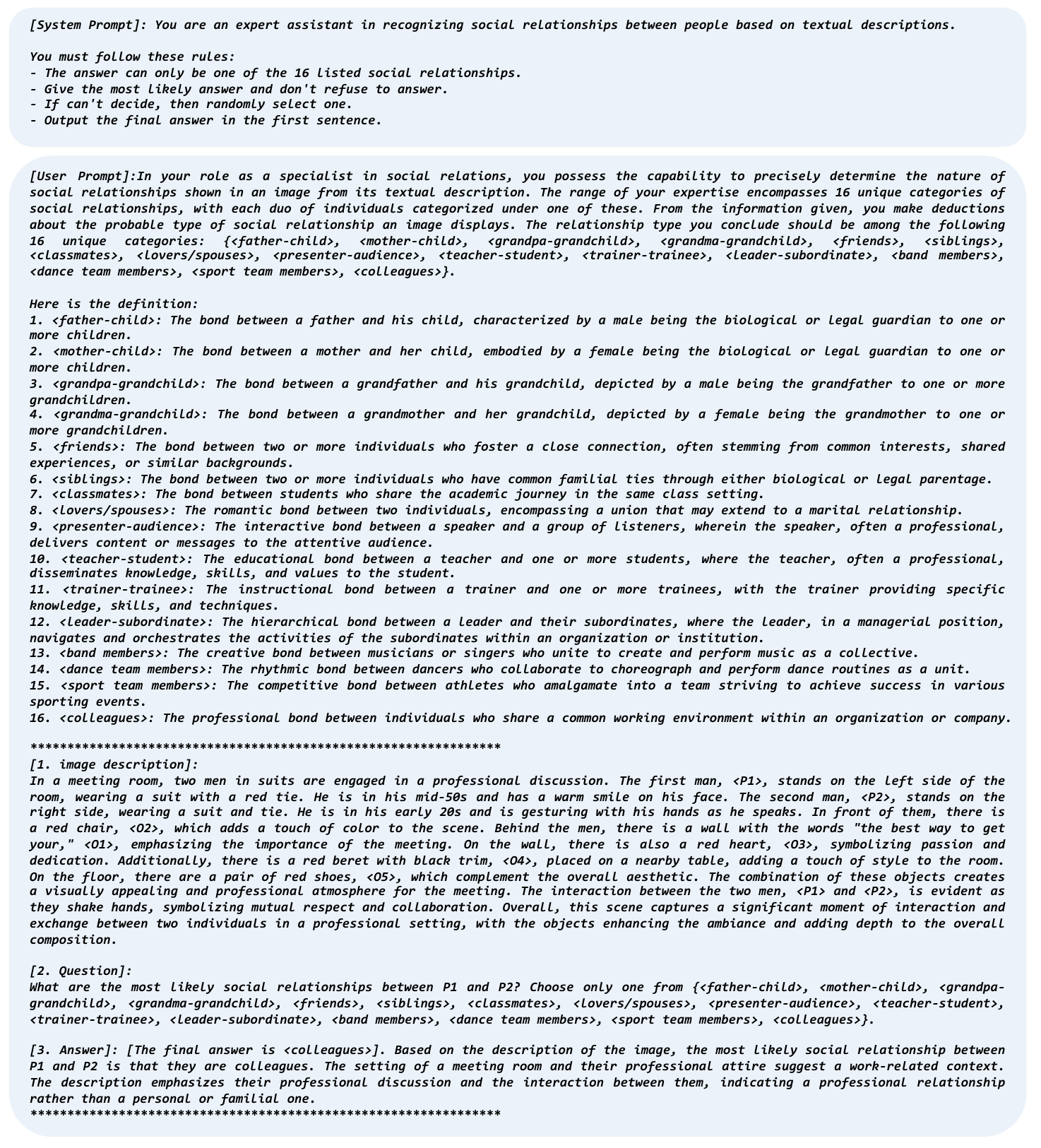}
    \caption{ The prompt after GSPO on the PIPA dataset.
    }  \label{fig:pipafinal}
\end{figure*}

\begin{figure*}[t]
    \centering
    \includegraphics[width=1.0\textwidth]{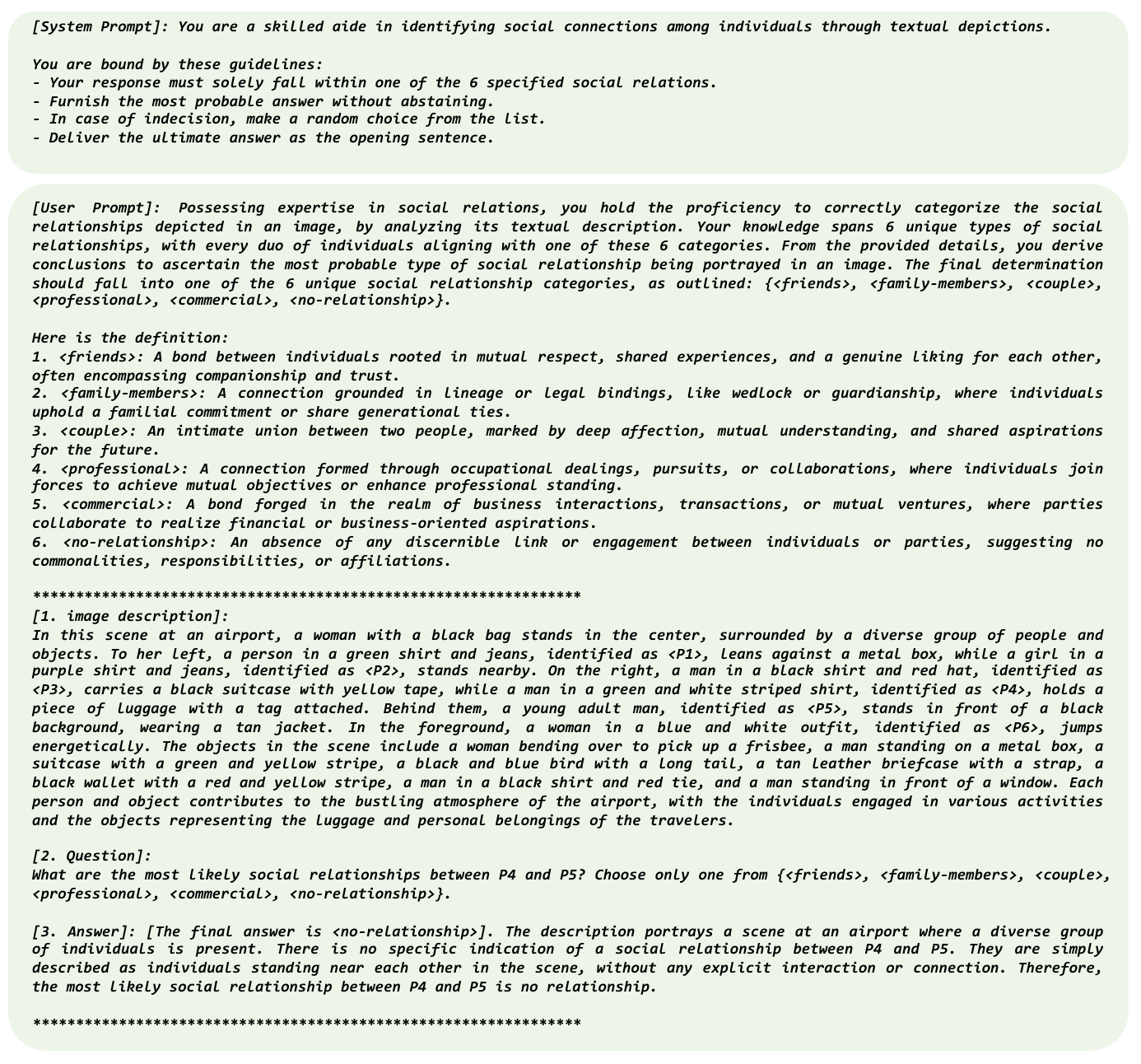}
    \caption{ The prompt after GSPO on the PISC dataset.
    }  \label{fig:piscfinal}
\end{figure*}

\noindent\textbf{{\namep} after GSPO.} Due to the time and effort-intensive nature of manually designing prompts, this paper introduces the Greedy Segment Prompt Optimization method. For each segment, we employ ChatGPT to generate multiple candidates. As for the in-context examples, we also randomly select several samples from the training dataset.
Here, we employ Vicuna-7B~\cite{vicuna2023} for training to obtain the optimized prompts. The optimized prompt on the PIPA dataset is illustrated in Figure \ref{fig:pipafinal}, while that on the PISC dataset is shown in Figure \ref{fig:piscfinal}.

\clearpage
%%%%%%%%%%%%%%%%%%%%%%%%%%%%%%%%%%%%%%%%%%%%%%%%%%%%%%%%%%%%

\newpage
\section*{NeurIPS Paper Checklist}

\begin{enumerate}

\item {\bf Claims}
    \item[] Question: Do the main claims made in the abstract and introduction accurately reflect the paper's contributions and scope?
    \item[] Answer:  \answerYes{} % Replace by \answerYes{}, \answerNo{}, or \answerNA{}.
    \item[] Justification: The main claims made in the abstract and introduction accurately reflect the paper's contributions and scope.
    \item[] Guidelines:
    \begin{itemize}
        \item The answer NA means that the abstract and introduction do not include the claims made in the paper.
        \item The abstract and/or introduction should clearly state the claims made, including the contributions made in the paper and important assumptions and limitations. A No or NA answer to this question will not be perceived well by the reviewers. 
        \item The claims made should match theoretical and experimental results, and reflect how much the results can be expected to generalize to other settings. 
        \item It is fine to include aspirational goals as motivation as long as it is clear that these goals are not attained by the paper. 
    \end{itemize}

\item {\bf Limitations}
    \item[] Question: Does the paper discuss the limitations of the work performed by the authors?
    \item[] Answer: \answerYes{} % Replace by \answerYes{}, \answerNo{}, or \answerNA{}.
    \item[] Justification: We discussed the limitations in the Conclusion Section.
    \item[] Guidelines: 
    \begin{itemize}
        \item The answer NA means that the paper has no limitation while the answer No means that the paper has limitations, but those are not discussed in the paper. 
        \item The authors are encouraged to create a separate "Limitations" section in their paper.
        \item The paper should point out any strong assumptions and how robust the results are to violations of these assumptions (e.g., independence assumptions, noiseless settings, model well-specification, asymptotic approximations only holding locally). The authors should reflect on how these assumptions might be violated in practice and what the implications would be.
        \item The authors should reflect on the scope of the claims made, e.g., if the approach was only tested on a few datasets or with a few runs. In general, empirical results often depend on implicit assumptions, which should be articulated.
        \item The authors should reflect on the factors that influence the performance of the approach. For example, a facial recognition algorithm may perform poorly when image resolution is low or images are taken in low lighting. Or a speech-to-text system might not be used reliably to provide closed captions for online lectures because it fails to handle technical jargon.
        \item The authors should discuss the computational efficiency of the proposed algorithms and how they scale with dataset size.
        \item If applicable, the authors should discuss possible limitations of their approach to address problems of privacy and fairness.
        \item While the authors might fear that complete honesty about limitations might be used by reviewers as grounds for rejection, a worse outcome might be that reviewers discover limitations that aren't acknowledged in the paper. The authors should use their best judgment and recognize that individual actions in favor of transparency play an important role in developing norms that preserve the integrity of the community. Reviewers will be specifically instructed to not penalize honesty concerning limitations.
    \end{itemize}

\item {\bf Theory Assumptions and Proofs}
    \item[] Question: For each theoretical result, does the paper provide the full set of assumptions and a complete (and correct) proof?
    \item[] Answer: \answerNA{} % Replace by \answerYes{}, \answerNo{}, or \answerNA{}.
    \item[] Justification: This paper does not involve theoretical contributions.
    \item[] Guidelines:
    \begin{itemize}
        \item The answer NA means that the paper does not include theoretical results. 
        \item All the theorems, formulas, and proofs in the paper should be numbered and cross-referenced.
        \item All assumptions should be clearly stated or referenced in the statement of any theorems.
        \item The proofs can either appear in the main paper or the supplemental material, but if they appear in the supplemental material, the authors are encouraged to provide a short proof sketch to provide intuition. 
        \item Inversely, any informal proof provided in the core of the paper should be complemented by formal proofs provided in appendix or supplemental material.
        \item Theorems and Lemmas that the proof relies upon should be properly referenced. 
    \end{itemize}

    \item {\bf Experimental Result Reproducibility}
    \item[] Question: Does the paper fully disclose all the information needed to reproduce the main experimental results of the paper to the extent that it affects the main claims and/or conclusions of the paper (regardless of whether the code and data are provided or not)?
    \item[] Answer: \answerYes{} % Replace by \answerYes{}, \answerNo{}, or \answerNA{}.
    \item[] Justification: We have included an Algorithm to clearly demonstrate how to reproduce our method. We will also release the code. 
    \item[] Guidelines:
    \begin{itemize}
        \item The answer NA means that the paper does not include experiments.
        \item If the paper includes experiments, a No answer to this question will not be perceived well by the reviewers: Making the paper reproducible is important, regardless of whether the code and data are provided or not.
        \item If the contribution is a dataset and/or model, the authors should describe the steps taken to make their results reproducible or verifiable. 
        \item Depending on the contribution, reproducibility can be accomplished in various ways. For example, if the contribution is a novel architecture, describing the architecture fully might suffice, or if the contribution is a specific model and empirical evaluation, it may be necessary to either make it possible for others to replicate the model with the same dataset, or provide access to the model. In general. releasing code and data is often one good way to accomplish this, but reproducibility can also be provided via detailed instructions for how to replicate the results, access to a hosted model (e.g., in the case of a large language model), releasing of a model checkpoint, or other means that are appropriate to the research performed.
        \item While NeurIPS does not require releasing code, the conference does require all submissions to provide some reasonable avenue for reproducibility, which may depend on the nature of the contribution. For example
        \begin{enumerate}
            \item If the contribution is primarily a new algorithm, the paper should make it clear how to reproduce that algorithm.
            \item If the contribution is primarily a new model architecture, the paper should describe the architecture clearly and fully.
            \item If the contribution is a new model (e.g., a large language model), then there should either be a way to access this model for reproducing the results or a way to reproduce the model (e.g., with an open-source dataset or instructions for how to construct the dataset).
            \item We recognize that reproducibility may be tricky in some cases, in which case authors are welcome to describe the particular way they provide for reproducibility. In the case of closed-source models, it may be that access to the model is limited in some way (e.g., to registered users), but it should be possible for other researchers to have some path to reproducing or verifying the results.
        \end{enumerate}
    \end{itemize}

\item {\bf Open access to data and code}
    \item[] Question: Does the paper provide open access to the data and code, with sufficient instructions to faithfully reproduce the main experimental results, as described in supplemental material?
    \item[] Answer: \answerYes{} % Replace by \answerYes{}, \answerNo{}, or \answerNA{}.
    \item[] Justification: The source codes will be made available to the public.
    \item[] Guidelines:
    \begin{itemize}
        \item The answer NA means that paper does not include experiments requiring code.
        \item Please see the NeurIPS code and data submission guidelines (\url{https://nips.cc/public/guides/CodeSubmissionPolicy}) for more details.
        \item While we encourage the release of code and data, we understand that this might not be possible, so “No” is an acceptable answer. Papers cannot be rejected simply for not including code, unless this is central to the contribution (e.g., for a new open-source benchmark).
        \item The instructions should contain the exact command and environment needed to run to reproduce the results. See the NeurIPS code and data submission guidelines (\url{https://nips.cc/public/guides/CodeSubmissionPolicy}) for more details.
        \item The authors should provide instructions on data access and preparation, including how to access the raw data, preprocessed data, intermediate data, and generated data, etc.
        \item The authors should provide scripts to reproduce all experimental results for the new proposed method and baselines. If only a subset of experiments are reproducible, they should state which ones are omitted from the script and why.
        \item At submission time, to preserve anonymity, the authors should release anonymized versions (if applicable).
        \item Providing as much information as possible in supplemental material (appended to the paper) is recommended, but including URLs to data and code is permitted.
    \end{itemize}

\item {\bf Experimental Setting/Details}
    \item[] Question: Does the paper specify all the training and test details (e.g., data splits, hyperparameters, how they were chosen, type of optimizer, etc.) necessary to understand the results?
    \item[] Answer: \answerYes{} % Replace by \answerYes{}, \answerNo{}, or \answerNA{}.
    \item[] Justification: We have included it in the Experiments Section.
    \item[] Guidelines:
    \begin{itemize}
        \item The answer NA means that the paper does not include experiments.
        \item The experimental setting should be presented in the core of the paper to a level of detail that is necessary to appreciate the results and make sense of them.
        \item The full details can be provided either with the code, in appendix, or as supplemental material.
    \end{itemize}

\item {\bf Experiment Statistical Significance}
    \item[] Question: Does the paper report error bars suitably and correctly defined or other appropriate information about the statistical significance of the experiments?
    \item[] Answer: \answerNA{} % Replace by \answerYes{}, \answerNo{}, or \answerNA{}.
    \item[] Justification: It is not included in all previous work in this field.
    \item[] Guidelines:
    \begin{itemize}
        \item The answer NA means that the paper does not include experiments.
        \item The authors should answer "Yes" if the results are accompanied by error bars, confidence intervals, or statistical significance tests, at least for the experiments that support the main claims of the paper.
        \item The factors of variability that the error bars are capturing should be clearly stated (for example, train/test split, initialization, random drawing of some parameter, or overall run with given experimental conditions).
        \item The method for calculating the error bars should be explained (closed form formula, call to a library function, bootstrap, etc.)
        \item The assumptions made should be given (e.g., Normally distributed errors).
        \item It should be clear whether the error bar is the standard deviation or the standard error of the mean.
        \item It is OK to report 1-sigma error bars, but one should state it. The authors should preferably report a 2-sigma error bar than state that they have a 96\% CI, if the hypothesis of Normality of errors is not verified.
        \item For asymmetric distributions, the authors should be careful not to show in tables or figures symmetric error bars that would yield results that are out of range (e.g. negative error rates).
        \item If error bars are reported in tables or plots, The authors should explain in the text how they were calculated and reference the corresponding figures or tables in the text.
    \end{itemize}

\item {\bf Experiments Compute Resources}
    \item[] Question: For each experiment, does the paper provide sufficient information on the computer resources (type of compute workers, memory, time of execution) needed to reproduce the experiments?
    \item[] Answer: \answerYes{} % Replace by \answerYes{}, \answerNo{}, or \answerNA{}.
    \item[] Justification: We have included it in the Experiments Section.
    \item[] Guidelines:
    \begin{itemize}
        \item The answer NA means that the paper does not include experiments.
        \item The paper should indicate the type of compute workers CPU or GPU, internal cluster, or cloud provider, including relevant memory and storage.
        \item The paper should provide the amount of compute required for each of the individual experimental runs as well as estimate the total compute. 
        \item The paper should disclose whether the full research project required more compute than the experiments reported in the paper (e.g., preliminary or failed experiments that didn't make it into the paper). 
    \end{itemize}
    
\item {\bf Code Of Ethics}
    \item[] Question: Does the research conducted in the paper conform, in every respect, with the NeurIPS Code of Ethics \url{https://neurips.cc/public/EthicsGuidelines}?
    \item[] Answer: \answerYes{} % Replace by \answerYes{}, \answerNo{}, or \answerNA{}.
    \item[] Justification:  Our research conducted in the paper conforms to the NeurIPS Code of Ethics.
    \item[] Guidelines:
    \begin{itemize}
        \item The answer NA means that the authors have not reviewed the NeurIPS Code of Ethics.
        \item If the authors answer No, they should explain the special circumstances that require a deviation from the Code of Ethics.
        \item The authors should make sure to preserve anonymity (e.g., if there is a special consideration due to laws or regulations in their jurisdiction).
    \end{itemize}

\item {\bf Broader Impacts}
    \item[] Question: Does the paper discuss both potential positive societal impacts and negative societal impacts of the work performed?
    \item[] Answer: \answerYes{} % Replace by \answerYes{}, \answerNo{}, or \answerNA{}.
    \item[] Justification: We have discussed the broader impact in the Conclusion Section.
    \item[] Guidelines:
    \begin{itemize}
        \item The answer NA means that there is no societal impact of the work performed.
        \item If the authors answer NA or No, they should explain why their work has no societal impact or why the paper does not address societal impact.
        \item Examples of negative societal impacts include potential malicious or unintended uses (e.g., disinformation, generating fake profiles, surveillance), fairness considerations (e.g., deployment of technologies that could make decisions that unfairly impact specific groups), privacy considerations, and security considerations.
        \item The conference expects that many papers will be foundational research and not tied to particular applications, let alone deployments. However, if there is a direct path to any negative applications, the authors should point it out. For example, it is legitimate to point out that an improvement in the quality of generative models could be used to generate deepfakes for disinformation. On the other hand, it is not needed to point out that a generic algorithm for optimizing neural networks could enable people to train models that generate Deepfakes faster.
        \item The authors should consider possible harms that could arise when the technology is being used as intended and functioning correctly, harms that could arise when the technology is being used as intended but gives incorrect results, and harms following from (intentional or unintentional) misuse of the technology.
        \item If there are negative societal impacts, the authors could also discuss possible mitigation strategies (e.g., gated release of models, providing defenses in addition to attacks, mechanisms for monitoring misuse, mechanisms to monitor how a system learns from feedback over time, improving the efficiency and accessibility of ML).
    \end{itemize}
    
\item {\bf Safeguards}
    \item[] Question: Does the paper describe safeguards that have been put in place for responsible release of data or models that have a high risk for misuse (e.g., pretrained language models, image generators, or scraped datasets)?
    \item[] Answer: \answerNA{} % Replace by \answerYes{}, \answerNo{}, or \answerNA{}.
    \item[] Justification: Our research doesn't train new models. We use open-sourced foundation models, and any safeguards they used can be applied to our method. 
    \item[] Guidelines:
    \begin{itemize}
        \item The answer NA means that the paper poses no such risks.
        \item Released models that have a high risk for misuse or dual-use should be released with necessary safeguards to allow for controlled use of the model, for example by requiring that users adhere to usage guidelines or restrictions to access the model or implementing safety filters. 
        \item Datasets that have been scraped from the Internet could pose safety risks. The authors should describe how they avoided releasing unsafe images.
        \item We recognize that providing effective safeguards is challenging, and many papers do not require this, but we encourage authors to take this into account and make a best faith effort.
    \end{itemize}

\item {\bf Licenses for existing assets}
    \item[] Question: Are the creators or original owners of assets (e.g., code, data, models), used in the paper, properly credited and are the license and terms of use explicitly mentioned and properly respected?
    \item[] Answer: \answerYes{} % Replace by \answerYes{}, \answerNo{}, or \answerNA{}.
    \item[] Justification: We have properly cited the original paper.
    \item[] Guidelines:
    \begin{itemize}
        \item The answer NA means that the paper does not use existing assets.
        \item The authors should cite the original paper that produced the code package or dataset.
        \item The authors should state which version of the asset is used and, if possible, include a URL.
        \item The name of the license (e.g., CC-BY 4.0) should be included for each asset.
        \item For scraped data from a particular source (e.g., website), the copyright and terms of service of that source should be provided.
        \item If assets are released, the license, copyright information, and terms of use in the package should be provided. For popular datasets, \url{paperswithcode.com/datasets} has curated licenses for some datasets. Their licensing guide can help determine the license of a dataset.
        \item For existing datasets that are re-packaged, both the original license and the license of the derived asset (if it has changed) should be provided.
        \item If this information is not available online, the authors are encouraged to reach out to the asset's creators.
    \end{itemize}

\item {\bf New Assets}
    \item[] Question: Are new assets introduced in the paper well documented and is the documentation provided alongside the assets?
    \item[] Answer: \answerNA{} % Replace by \answerYes{}, \answerNo{}, or \answerNA{}.
    \item[] Justification: This paper does not release new assets
    \item[] Guidelines:
    \begin{itemize}
        \item The answer NA means that the paper does not release new assets.
        \item Researchers should communicate the details of the dataset/code/model as part of their submissions via structured templates. This includes details about training, license, limitations, etc. 
        \item The paper should discuss whether and how consent was obtained from people whose asset is used.
        \item At submission time, remember to anonymize your assets (if applicable). You can either create an anonymized URL or include an anonymized zip file.
    \end{itemize}

\item {\bf Crowdsourcing and Research with Human Subjects}
    \item[] Question: For crowdsourcing experiments and research with human subjects, does the paper include the full text of instructions given to participants and screenshots, if applicable, as well as details about compensation (if any)? 
    \item[] Answer: \answerNA{} % Replace by \answerYes{}, \answerNo{}, or \answerNA{}.
    \item[] Justification: The paper does not involve crowdsourcing nor research with human subjects.
    \item[] Guidelines:
    \begin{itemize}
        \item The answer NA means that the paper does not involve crowdsourcing nor research with human subjects.
        \item Including this information in the supplemental material is fine, but if the main contribution of the paper involves human subjects, then as much detail as possible should be included in the main paper. 
        \item According to the NeurIPS Code of Ethics, workers involved in data collection, curation, or other labor should be paid at least the minimum wage in the country of the data collector. 
    \end{itemize}

\item {\bf Institutional Review Board (IRB) Approvals or Equivalent for Research with Human Subjects}
    \item[] Question: Does the paper describe potential risks incurred by study participants, whether such risks were disclosed to the subjects, and whether Institutional Review Board (IRB) approvals (or an equivalent approval/review based on the requirements of your country or institution) were obtained?
    \item[] Answer: \answerNA{} % Replace by \answerYes{}, \answerNo{}, or \answerNA{}.
    \item[] Justification: The paper does not involve crowdsourcing nor research with human subjects.
    \item[] Guidelines:
    \begin{itemize}
        \item The answer NA means that the paper does not involve crowdsourcing nor research with human subjects.
        \item Depending on the country in which research is conducted, IRB approval (or equivalent) may be required for any human subjects research. If you obtained IRB approval, you should clearly state this in the paper. 
        \item We recognize that the procedures for this may vary significantly between institutions and locations, and we expect authors to adhere to the NeurIPS Code of Ethics and the guidelines for their institution. 
        \item For initial submissions, do not include any information that would break anonymity (if applicable), such as the institution conducting the review.
    \end{itemize}

\end{enumerate}

\end{document}